\documentclass[]{amsart}

\newcommand{\xadded}[1]{#1}
\newcommand{\xdeleted}[1]{}
\newcommand{\xreplaced}[2]{#1}

\usepackage{_packages}
\usepackage{_macros}
\usepackage{authblk}

\title{Towards Single Camera Human 3D-Kinematics }

\newcommand{\vv}{\quad Vicarious Perception Technologies (VicarVision), 1015 AH Amsterdam, The Netherlands}
\newcommand{\tudcv}{\quad Computer Vision Lab, Delft University of Technology, 2628 XE Delft, The Netherlands}
\newcommand{\tudbme}{\quad Biomechanical Engineering, Delft University of Technology, 2628 CN Delft, The Netherlands}

\usepackage[foot]{amsaddr}
\author{Marian Bittner %MDPI: 1. Please carefully check the accuracy of names and affiliations. 2. contributed equally author should be two authors, please check if the \dagger can be removed here or added \dagger for another author 
 $^{1,2,3,}$*$^{,\dagger}$ \and Wei-Tse Yang $^{2,\dagger}$ \and Xucong Zhang $^{2}$ \and Ajay Seth $^{3}$ \and Jan van Gemert $^{2}$ \and Frans C. T. van der Helm $^{3}$}

\address{$^1$\vv}
\address{$^2$\tudcv}
\address{$^3$\tudbme}
\address{*Corresponding author: mbittner.work@gmail.com}
\address{$\dagger$These authors contributed equally to this work.}

\begin{document}

\maketitle

\begin{abstract}
Markerless estimation of 3D Kinematics has the great potential to clinically diagnose and monitor movement disorders without referrals to expensive motion capture labs; however, current approaches are limited by performing multiple de-coupled steps to estimate the kinematics of a person from videos. 
Most current techniques work in a multi-step approach by first detecting the pose of the body and then fitting a musculoskeletal model to the data for accurate kinematic estimation. Errors in training data of the pose detection algorithms, model scaling, as well the requirement of multiple cameras limit the use of these techniques in a clinical setting. 
Our goal is to pave the way toward fast, easily applicable and accurate 3D kinematic estimation \xdeleted{in a clinical setting}. To this end, we propose a novel approach for direct 3D human kinematic estimation D3KE
%MDPI: Please confirm if the bold is unnecessary and can be removed. The following highlights are the same. %MBittner: Confirmed and removed.
 from videos using deep neural networks. Our experiments demonstrate that the proposed end-to-end training is robust and outperforms 2D and 3D markerless motion capture based kinematic estimation pipelines in terms of joint angles error by a large margin (35\% from 5.44 to 3.54 degrees). We show that D3KE is superior to the multi-step approach and can run at video framerate speeds.
This technology shows the potential for clinical analysis from mobile devices in the future.
\end{abstract}

% The fields PACS, MSC, and JEL may be left empty or commented out if not applicable
%\PACS{J0101}
%\MSC{}
%\JEL{}

%%%%%%%%%%%%%%%%%%%%%%%%%%%%%%%%%%%%%%%%%%

% \input{1_introduction}
%\input{1_introduction_n}
\section{Introduction}

\xadded{
3D Human kinematics %MDPI: we removed the comment before this para, please confirm % MBittner: Comment was only left for clarity during the review process. Confirmed.
 involves measuring joint angles between body segments, which is essential in the day-to-day practice of experts.
Skilled physicians could judge, just by looking at a specific motion of their patient, whether it is healthy or abnormal. Skilled sports coaches can help their coachees achieve better performance and lower injury risk by evaluating their movements through observation. However, these visual examinations of human kinematics remain inherently subjective, leading to variation between and within human observers. Modern systems and sensors could reduce these variations through more objective observations. Yet, these systems make the measurement of human motion more costly and more time-consuming. A~system with the availability and ease of use of visual estimation would help physicians and coaches make more objective observations more often, ultimately raising their own and their subjects quality of life.}
\xadded{ 
Digital cameras have made the estimation of human kinematics more accessible but come at the cost of reduced accuracy. Compared to the more traditional Optical Motion capture (OMC) systems, markerless motion capture  (MMC) systems do not require specialized cameras and markers attached to the subject being monitored, but~use normal RGB cameras in combination with image-based automatic pose estimation algorithms. Instead of specific markers, pose estimation algorithms detect the centers of major joints of the human body, such as the shoulders, hips, and~knees. These detected centers are usually referred to as key points.}

\xadded{Multiple commonly used markerless motion capture methods rely on 2D pose estimation methods~\cite{kidzinski_deep_2020,pagnon_pose2sim_2021,pagnon_pose2sim_2022,kanko_assessment_2021}. Often these methods still need more than one camera to generate a good estimation of the keypoints in 3D, which again requires additional cameras to be set up. On~the other hand, an~increasing number of methods  are using single-view (monocular) 3D pose estimation methods~\cite{gu_markerless_2018,liao_model-based_2020,noteboom_feasibility_2022}, which allow to estimate a 3D pose just by using a single camera.
This makes MMC systems faster and more accessible as they do not require the additional time and expertise to place markers on the subject or calibrate multiple cameras.
However, MMC systems assume that current pose estimation algorithms can accurately replace markerless motion capture systems for, e.g.,~biomechanical applications~\cite{seethapathi_movement_2019,cronin_using_2021,wade_applications_2022}. }

{
Commonly used pose estimation algorithms introduce mistakes in kinematic estimation pipelines due to systematic errors in their predictions. To~detect key points, most pose estimation methods are trained on a combination of images of a person and ground truth annotations which map pixels in the image to their corresponding joint center.These ground truth annotations are often manually conducted by non-expert annotators, leading to errors caused by personal biases for training and inaccuracies in the pose estimations~\cite{cronin_using_2021}. For~example, Needham~et~al.~\cite{needham_accuracy_2021} compared three often used pose estimation algorithms OpenPose~\cite{OpenPose}, DeepLabCut~\cite{mathis_deeplabcut_2018} and AlphaPose~\cite{fang_rmpe_2017} algorithm against an OMC system and showed errors in the estimation of joint centers of 30 mm to 50 mm with variations in 12 mm to 25 mm in marker placement. Cronin~\cite{cronin_using_2021} provides an overview of additional problems with 2D pose estimation for kinematic analysis. 
These differences are most likely due to a difference between the application that pose estimation algorithms are often developed for and their application to, e.g.,~the biomedical domain, which has different accuracy requirements~\cite{seethapathi_movement_2019}.
Wade~et~al.~\cite{wade_applications_2022} proposed to solve this problem by re-annotating existing large-scale datasets, this, however, is a time-consuming process, when for example considering the COCO-keypoint dataset \url{https://cocodataset.org/#keypoints-2020} (accessed on 2 December 2022 %MDPI: Footnote is not permitted in our journal. Please include this paragraph in the maintext.
 %MDPI: Please provide the access date of the URL in the following format: "URL (accessed on Day Month Year)". % MBittner: Done. Added 'accessed on 2 December 2022'.
) consists of more than 250.000 labeled poses.
For the evaluation of pose estimation algorithms, these labeling errors will just appear as a baseline error that all algorithms training on the same data will have.
However, for~applications in the biomedical domain and in~situations such as kinematic estimation, where the pose is just an intermediate step errors can propagate to subsequent tasks. 
}

\xadded{
Errors in the estimated pose cannot not be corrected by most kinematic estimation pipelines because they all roughly follow a `multi-step' approach.
The `multi-step' approach consists of}
\xadded{
\begin{itemize}
    \item Detection of the 3D pose (in one or more steps);
    \item (Optional) modeling of the pose with a (musculo)skeletal model. 
    \item Calculation of kinematics and/or downstream tasks such as gait parameters or dynamics.
\end{itemize}
}

\xadded{ For example, Kidzinski~et~al.~\cite{kidzinski_deep_2020} used OpenPose to first predict key points from a video and then trained a convolutional neural network (CNN) to predict the walking parameters of patients with cerebral palsy. Liao~et~al.~\cite{liao_model-based_2020} first model the 2D pose in OpenPose then create a 3D pose using data-driven matching and finally estimate 3D gait parameters.
Noteboom~et~al.~\cite{noteboom_feasibility_2022} first used VideoPose3D~\cite{pavllo:videopose3d:2019} to estimate a 3D pose, followed by modeling in OpenSim~\cite{seth_opensim_2018} for the estimation of dynamics from a single camera.
\mbox{Pagnon~et~al.}~\cite{pagnon_pose2sim_2021,pagnon_pose2sim_2022} developed the handy Pose2Sim tool, which first combines 2D OpenPose pose estimations from multiple cameras into a 3D pose then models it in OpenSim.
% \jvg{Do you mean: "Because the pose estimation steps is  de-coupled from kinematic estimation,"} \mb{Yes!} 
Because the pose estimation step is de-coupled from kinematic estimation, errors in pose estimation propagate through to the estimation of kinematics.
Uchida and Seth~\cite{uchida_conclusion_2022} showed that \SI{20}{\milli\meter} of marker uncertainty leads to a variation of 15.9$^{\circ}$ in peak ankle plantarflexion angle and impacts downstream tasks such as joint moment estimation.  
% \jvg{try to avoid starting a sentence with a number} 
\mbox{Della Croce~et~al.}~\cite{della_croce_pelvis_1999} showed precision variation 13 mm to 25 mm, which leads to differences in estimated joint angles up to \SI{10}{\degree}. Fonseca~et~al.~\cite{fonseca_conventional_2022} showed that misplacement of markers up to \SI{10}{\milli\meter} can lead to errors of 7$^{\circ}$ depending on the marker. 
With estimation errors of 30 mm to 50 mm in keypoint estimation~\cite{needham_accuracy_2021}, it is to be expected that these errors will substantially influence kinematic estimation from markerless motion capture. Low-pass filter~\cite{needham_can_2021,pagnon_pose2sim_2021} or bi-directional Kalman-filter~\cite{needham_can_2021} has been applied to compensate for noisy key point estimations, but~cannot correct for faults in keypoint detection. Subsequent modeling and kinematic calculation steps can only compensate for these inaccuracies.
This `multi-step' approach is probably inspired by the steps of a traditional OMC method, as~in the traditional OMC systems the pose detection step is done using a different system and is thus isolated from the other steps. In~camera-based kinematic estimation pipelines, however, the~de-coupling of individual steps is no longer necessary.}

\xadded{
Deep neural networks have often demonstrated their ability to outperform multi-step systems, by~implicitly learning individual steps through end-to-end training between an input and the desired output~\cite{simonyan_very_2015,krizhevsky_imagenet_2017}.
The main strength of deep neural networks lies in their ability to break down a highly complex task, in~this case, the~estimation of kinematics from videos, into~a sequence of simpler tasks, without~the need for intermediate `hand-crafted' representations~\cite{lecun_deep_2015,allen-zhu_backward_2021}.
Due to the fully differentiable nature of neural networks, it means that an error in estimation during training can influence all stages of the network and adjust them accordingly~\cite{allen-zhu_backward_2021}. This allows deep neural networks to directly estimate kinematics.
}

\begin{figure}[H]
%    \centering
    \includegraphics[width=\columnwidth]{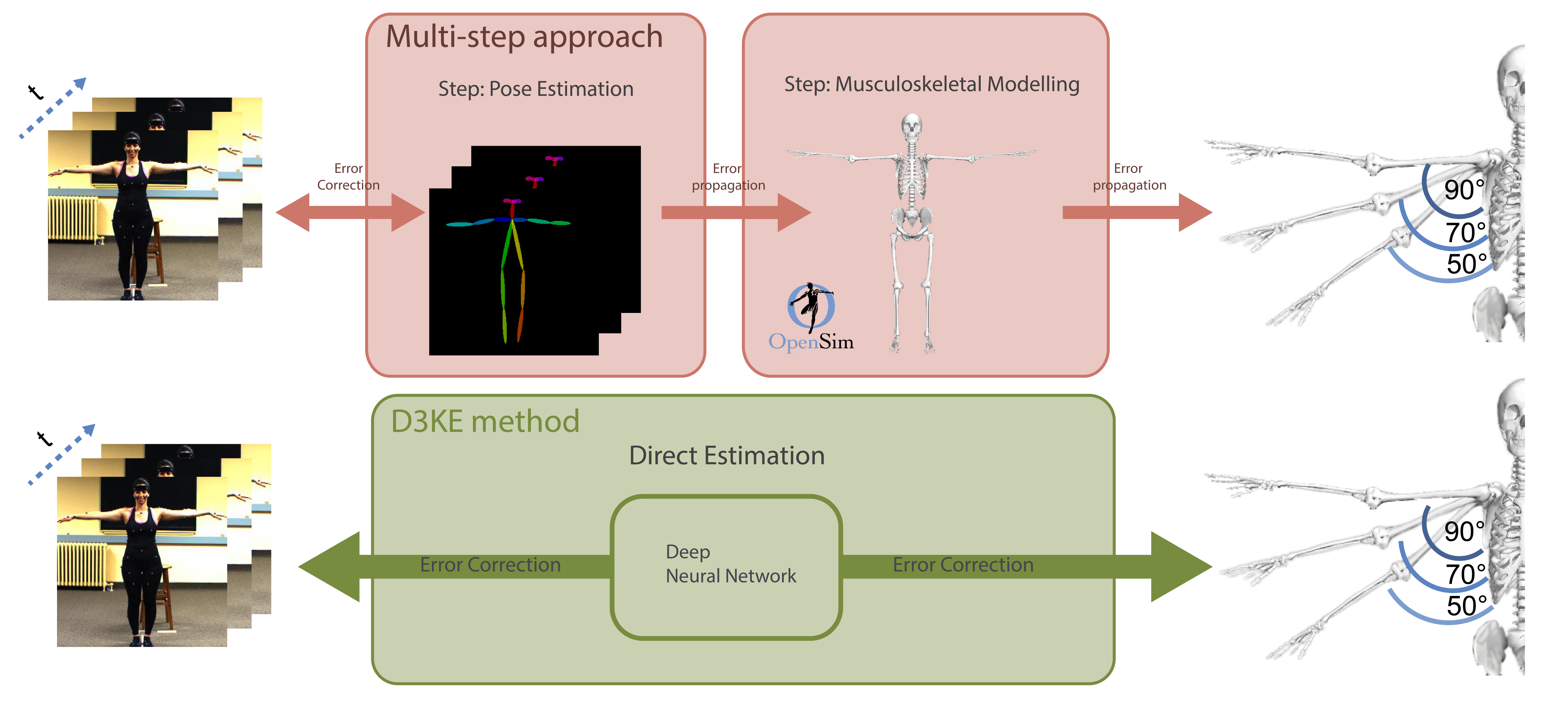}
    \caption{    
    Overview of the proposed direct 3D human kinematics estimation (D3KE). Instead of using the common 'multi-step' approach of predicting pose, fitting it to a model, and~estimating kinematics, our D3KE directly estimates the kinematics\xadded{. Errors in earlier steps of the multi-step approach propagate to later steps; in~contrast, our method can correct for errors occurring anywhere between input and output.}}
    \label{fig:paper_overview}
\end{figure}

In this work, we challenge the notion of the classical multi-step approach of pose estimation, fitting of a musculoskeletal model and kinematic estimation. To~this end, we propose a novel end-to-end method that allows for direct estimation of human kinematics, which is directly optimized for kinematic estimation while treating pose estimation only as an auxiliary task to constrain the estimations of the network. 
Figure~\ref{fig:paper_overview} shows a general overview of our~method.

\unskip

\subsection*{Contributions}

To the best of our knowledge, we are the first to present an end-to-end trainable network that directly generates joint angles, joint positions, scale factors and marker positions of a biomechanical model from a monocular video. We propose a method that directly regresses from a video \xdeleted{frame} 
to joint angles and scales using deep neural networks. We investigate the influence of 
\xreplaced{various}{different}
temporal smoothing methods to increase the accuracy of our algorithm. We introduce a novel type of network layer that allows for the calculation 
of the 3D pose from estimated kinematics during the training process to train the network simultaneously on the pose and kinematic~labels.

\section{Materials and Methods}

\begin{figure}[H]
%    \centering
  
%\begin{adjustwidth}{-\extralength}{0cm}
%\centering %% If there is a figure in wide page, please release command \centering
  \includegraphics[width=.98\columnwidth]{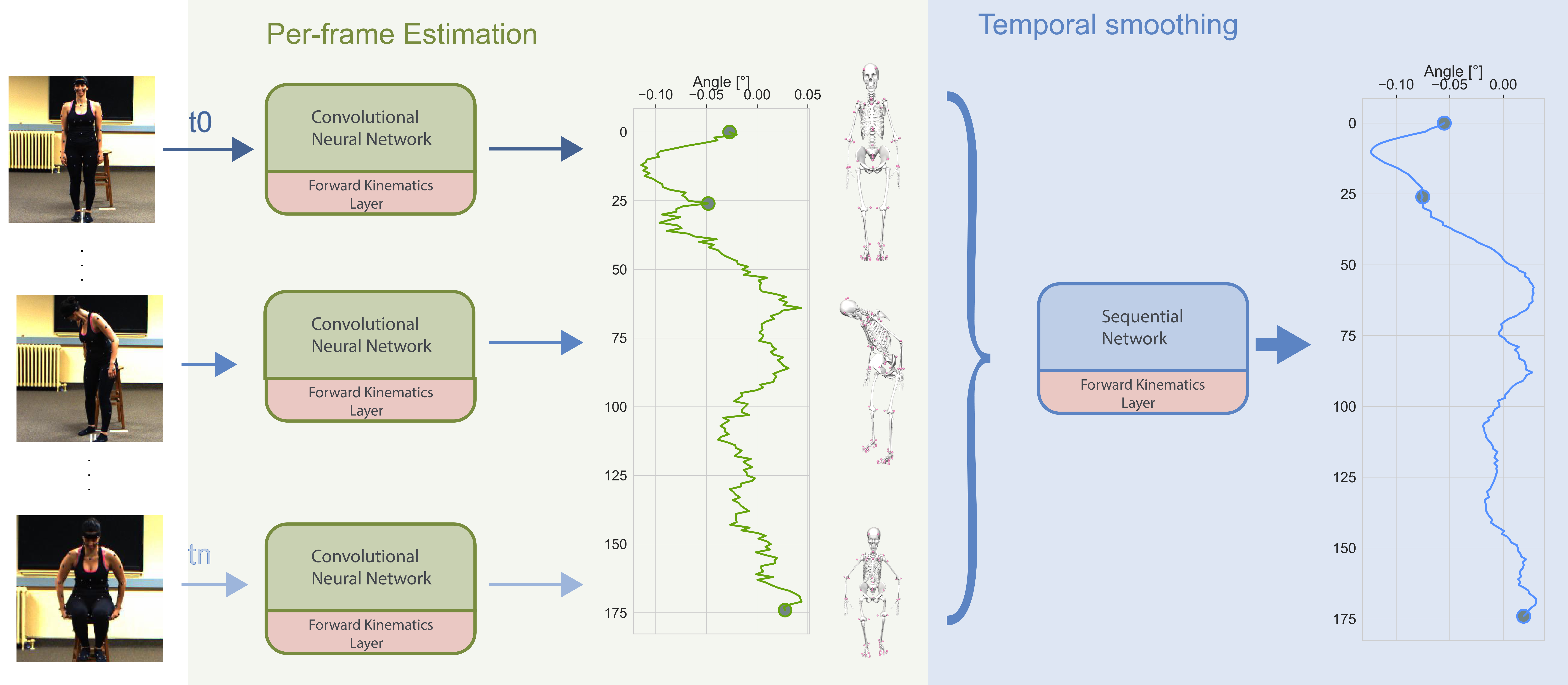}
%\end{adjustwidth}
    \caption{Taking a single view video as input, D3KE consists of one convolutional neural network and one sequential network. Per frame, D3KE outputs joint angles and scales of individual bones in a skeletal model(scale factors) with a convolutional network. Additionally, joint angle and scale factor are converted to a pose through the skeletal-model kinematics (SM) layer. A~series of frame estimations in time are then fed into a sequential network to smooth the estimations and reduce artifacts if one limb occludes another in the view of the camera (self-occlusion).    
    }
    \label{fig:method_overview}
\end{figure}

Our method takes videos from a single camera as input and directly estimates joint angles, \xreplaced{which}{and we} we call \xdeleted{it} direct 3D kinematic estimation (D3KE).\xdeleted{First,} The \xadded{proposed} method \xadded{first} coarsely \xreplaced{estimates kinematics per frame by using a convolutional neural network, and~then it uses a sequence network with temporal relations across frames to re-fine kinematic estimations at each frame}{predicts the musculoskeletal model parameters per frame by inferring body scales, joint angles and a rotation matrix from the pelvis to the ground using a convolutional neural network}. \xadded{ An overview of our method is shown in Figure~\ref{fig:method_overview}.
Both networks estimate the scale of body segments, joint angles, and~a rotation matrix from the pelvis to the ground, those serve as input for a skeletal-model layer in both networks that allows for additional supervision on the pose of a subject.}

%\unskip

In this section, \xadded{we first describe the deep learning architecture, including a detailed description of the skeletal-model layer. We then describe how the ground truth data was generated and which pre-processing and hyperparameters were used for training. Lastly, we describe the dataset used for training and testing our method.}

\subsection{Network~Structure}

% \jvg{"The proposed" = "Our"}\mb{Adressed}
\xadded{Convolutional neural networks(CNNs) have shown good accuracy for 2D and 3D pose estimations~\cite{OpenPose,MetricScale,VNect, Marginal_Heatmaps,coarsetofine} from single input images.
Conventionally 2D CNNs are used for pose estimation tasks, that takes a single image as an input and predict the pose of one or multiple people in the image.
For our method, we choose a per-frame convolutional network to coarsely predict the joint angle and scaling parameters.}
Inspired by~\cite{MetricScale}, we choose a standard pre-trained ResNeXt-50~\cite{ResNeXt} as our convolutional~backbone.

\xadded{To fine-tune the per-frame predicted joint angles and scaling parameters we add a sequential network.
Sequential networks are used in pose estimation to `lift' an estimated 2D pose to 3D~\cite{SpatialTemproal,occlusionAware}. Recent research combines temporal information with lifting to improve accuracy during frames where one limb occludes another in the view of the camera (self-occlusion) or where not all key points were detected~\cite{Transformer, TCN_ATTENION, pavllo:videopose3d:2019}. In~contrast to CNNs, these sequential networks do not take a single frame as input but exploit temporal dependencies in the data for their prediction. As~the convolutional network outputs per-frame estimates, it cannot take temporal information into account. We add a sequential network to our architecture to refine a sequence of estimations made by the convolutional model.
Inspired by works on temporal lifting we experimentally evaluate three sequential networks; an LSTM~\cite{LSTM}, a~Temporal Convolutional Network (TCN) \cite{pavllo:videopose3d:2019} and a Transformer~\cite{Transformer} to refine the predicted joint angles and scale~factors.

Both the convolutional and the sequential network contain a specialized layer that allows each network to perform the kinematic transformations of a musculo-skeletal model. Therefore, at~train time both networks can be supervised not only on the estimated joint angles but also on a resulting pose.}

Both convolutional and sequential networks are supervised by losses of joint positions, marker positions, body scales and joint angles. The~overall objective function $L$ can be expressed in the equation
\begin{equation} \label{lossFunc}
    L = \lambda_{1} L_{joint} + \lambda_{2} L_{marker} + \lambda_{3} L_{body} +\lambda_{4} L_{angle},
\end{equation}
where $\lambda_{1},\lambda_{2},\lambda_{3},\lambda_{4}$ are weights of losses.
We use the root-relative L1 loss in Equation~(\ref{rootRelative}) to define the loss of marker position $L_{marker}$ and the loss of joint position $L_{joint}$. The~estimations $\hat{y}$ and the labels $y$ are first subtracted with each root position $\hat{y}_{root}, y_{root}$. For~the loss of body scales $L_{body}$ and joint angles $L_{angle}$, we calculate the L1 norm.
\begin{equation} \label{rootRelative}
    l = \|(\hat{y} - \hat{y}_{root})- (y -  y_{root})\|_{1}
\end{equation}

\xadded{
The objective of a neural network during training is to minimize the loss function; in~our case, the difference between estimated and ground truth joint angles. 
However, this joint angle loss cannot capture the underlying relations and constraints of individual angles, dictated by the human musculoskeletal system. Intuitively, small changes in the angles of spine, shoulder, and~elbow can accumulate and lead to large differences in the position of the hand, as~illustrated in Figure~\ref{fig:fk_layer}. 
To address this issue, we propose to use a skeletal-model layer to perform the kinematic transform of a musculoskeletal~model.

\begin{figure}[H]
%    \centering
    \includegraphics[width=\columnwidth]{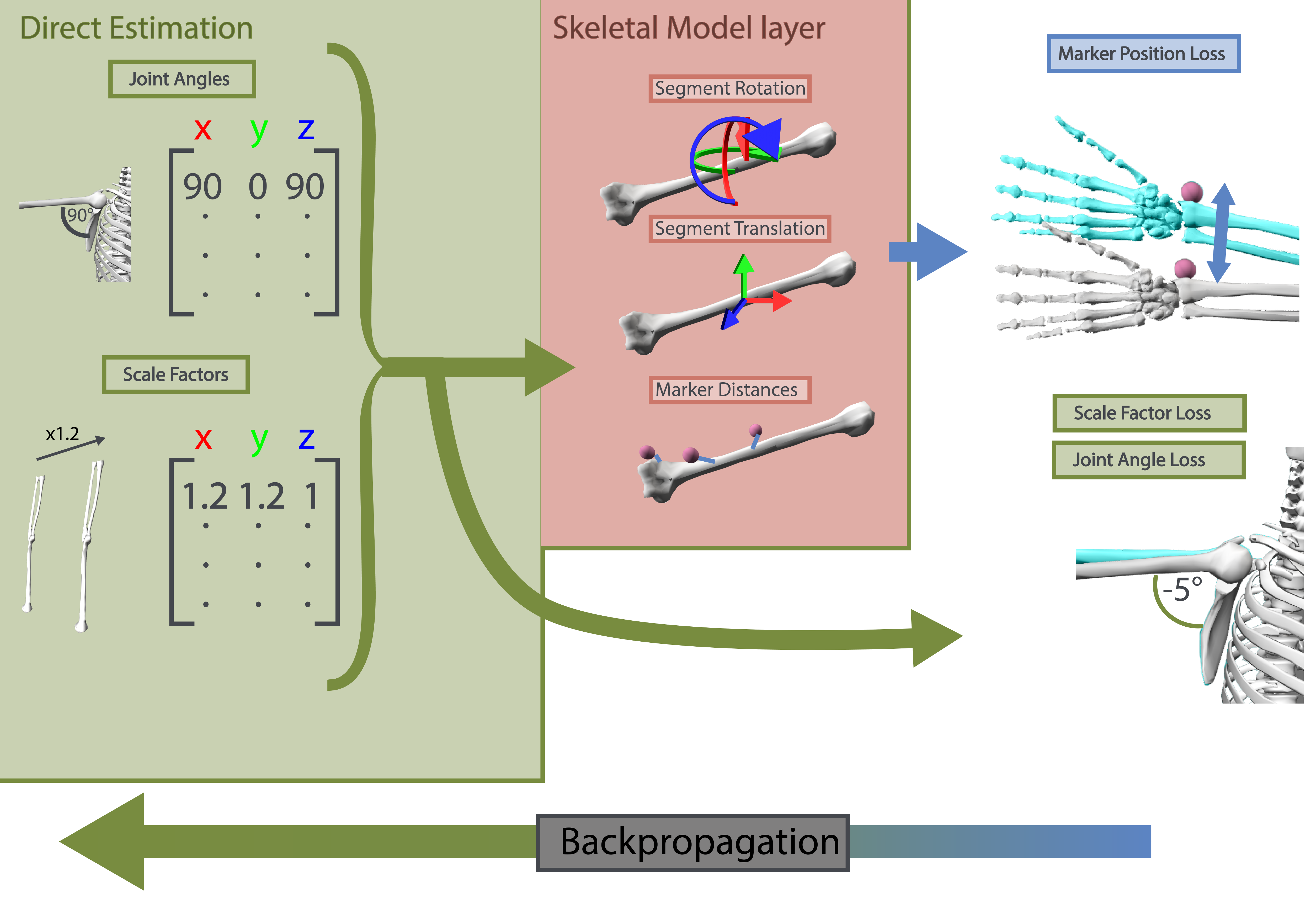}
    \caption{Our skeletal-model %MDPI: Please change the hyphen (-) into a minus sign (−, “U+2212”), e.g., “-5” should be “−5”. %MBittner: Done.
 layer uses an internal representation of a skeletal model to convert the  predicted joint angles and scale factors to the positions of individual markers on segments of the skeletal  model. This allows our method to be supervised during training not only on errors(losses) in the estimation of joint angles but also on errors in the resulting pose. On~the right, we show the additional error that is  created between estimations (gray) and ground truth (blue). This auxiliary estimation of the pose as 3D marker positions helps to constrain the estimation of joint angles as small changes in proximal joints can have a large effect on a marker at more distal~positions.
    }
    
    \label{fig:fk_layer}
\end{figure}

%{\captionof*{figure}{ }}
%\vspace{6pt}

\subsubsection*{Skeletal-Model~Layer}

The \xadded{skeletal-}model layer allows us to convert predicted joint angles into marker positions on a skeletal model and add them as an additional loss term. This loss term represents the cumulative effect of small joint angle changes on the final pose, indirectly imposing the constraints of a \xadded{skeletal-}model on the predictions of the network. As~the \xadded{skeletal-}model layer does not contain any learnable parameters, i.e.,~it cannot change during network training. The~accuracy of the predicted pose is completely determined by the input to the \xadded{skeletal-}model layer; thus, the pose prediction is only an auxiliary task.
}

A skeletal model consists of body segments, motions between different body segments (joints) and points with a vector from a center of its anchor body segment (markers). Given body scales $\beta$, joint angles $\theta$ and rotation matrix $R_{ground \leftarrow pelvis}$, we use the \xadded{skeletal-}model layer to calculate marker positions and joint positions. In~the following variables with a hat ($\hat{x}$) denote estimated values, variables without the hat ($x$) denote the predefined variable from the musculoskeletal~model.

First, the~translation part $T$ in the transformation from the joint to the body depends on the subject's body scale. For~example if the subject has longer legs, the~center of the femur will be farther from the hip joint. We can update the translation part $\hat{T}$ by comparing the ratio between predicted body scales $\hat{\beta}$ and default body scales $\beta$ in Equation~(\ref{updateTranslation}), where $\odot$ is elementwise multiplication, and~$\oslash$ is elementwise division.
\begin{equation} \label{updateTranslation}
    \hat{T} = T \odot (\hat{\beta} \oslash \beta)\\
\end{equation}

Then, we create a matrix to represent spatial transformation of motions $R_motion$ using Equation~(\ref{TransformRMotion}) $A_1,A_2,A_3$ are the predefined axes $\hat{\theta}_1,\hat{\theta}_2,\hat{\theta}_3$ and predicted angels per degree of freedom per joint in axis-angle notation. $G(A,\theta)$ is the standard function converting an axis-angle representation to a 3x3 transformation matrix.
\begin{equation} \label{TransformRMotion}
    R_{motion} = \begin{bmatrix}
 &  &  &  \\
 & R_3R_2R_1 &  & 0 \\
 &   &  &  \\
 & 0 &   & 1 \\
\end{bmatrix}_{4 \times 4}	
\end{equation}
\begin{equation} 
    R_1 = G(A_1,\theta_1)\\
\end{equation}
\begin{equation} \label{theta2}
    R_2 = G(R_1 A_2,\theta_2)\\
\end{equation}
\begin{equation} 
    R_3 = G(R_2R_1A_3,\theta_3)\\
\end{equation}

Then, we can calculate the estimated transformation from the body to its parent body $\hat{R}_{parent \leftarrow child}$ in Equation~(\ref{TransformChildParent}) using Equation~(\ref{TransformParentJoint}) with $O_{parent},O_{child}$ denoting predefined  orientations from and $ \hat{T}_{parent},\hat{T}_{child}$ the predicted translations from the joint to the parent/child. $F(O_{a})$ the conversion from euler angles to a 3 $\times$ 3 Rotation matrix.
\begin{equation} \label{TransformParentJoint}
    R_{parent/child\leftarrow joint}(O_{parent/child},T_{parent/child}) = \begin{bmatrix}
 &  &  &  \\
 & F(O_{parent/child}) &  & T_{parent/child} \\
 &   &  &  \\
 & 0 &   & 1 \\
\end{bmatrix}_{4\times4}	
\end{equation}
\begin{equation} \label{TransformChildParent}
    R_{parent\leftarrow child} = R_{parent\leftarrow joint}\  R_{motion} \ R_{child\leftarrow joint}^{-1}
\end{equation}

We measure the spatial transform by traversing from the root (pelvis) to leaf nodes (hands and feet) in the level order. 
In D3KE, we directly infer the rotation matrix from the pelvis to the ground $R_{ground \leftarrow pelvis}$. $R_{ground \leftarrow pelvis}$ can initially be expressed in Equation~(\ref{groundMatrix}), where $I$ denotes the identity matrix. The~rotation part of $R_{child \leftarrow joint}$ is also a 3 $\times$ 3 identity matrix in our musculoskeletal model. Our method aims to predict the root-relative position so the translation part can be ignored during the prediction. Moreover, in~our musculoskeletal model, only joint angles of the pelvis are unbounded in [$-\infty$,$\infty$]. Predicting three unbounded angles to form the rotation matrix in Equation~(\ref{TransformRMotion}) will have the problem of discontinuity~\cite{rotationContinuity}. Thus, we directly predict the rotation matrix $R_{ground \leftarrow pelvis}$.
\begin{equation} \label{groundMatrix}
    R_{ground\leftarrow pelvis} = I_{4 \times 4}\  R_{motion} \ R_{pelvis\leftarrow joint}^{-1}
\end{equation}

Last, a~marker with a vector of $\vec{d}$ from the center of the body is also dependent on the body scales. The~predicted vector of $\hat{d}$ is updated in Equation~(\ref{updateMarkerVector}).  The~position of the predicted point is calculated in Equation~(\ref{MarkerPosition}) with $\hat{R}_{parent \leftarrow child}$ and $\hat{d}$.
\begin{equation} \label{updateMarkerVector}
    \hat{d} = \vec{d} \odot (\hat{\beta} \oslash \beta)\\
\end{equation}
\begin{equation}\label{MarkerPosition}
    P = \prod_{parent,child\in path} \ R_{parent \leftarrow child}\begin{bmatrix}
 \\
 \vec{d}\\
 \\
 1
\end{bmatrix}_{4x1}
\end{equation}

\subsection{Network~Training}
\unskip
\subsubsection{Ground Truth~Generation }

\xadded{For training our method, we need to create custom ground truth data that contains all outcomes that our network is predicting since they are not available in publicly available datasets. 
Most pose estimation datasets, provide only video and marker positions from optical motion capture (OMC) system. For~training our method, we need the joint angle and the scales of individual bones, a~rotation matrix of the pelvis to the ground as well as the marker positions corresponding to them. 
To generate these, we model the OMC data, represented as a 3D human mesh model in the OpenSim software~\cite{seth_opensim_2018} and use inverse kinematics~\cite{lu_bone_1999, al_borno_opensense_2022} to generate joint angles. The~following describes each step in more detail.}

\xadded{First, we create a general (musculo)skeletal model to fit the data using the OpenSim software~\cite{seth_opensim_2018,OpenSim1}.}
As we are interested in capturing the complete motion of the human subject, we model the full body. With~the OpenSim software~\cite{OpenSim1, seth_opensim_2018} we create a full-body musculoskeletal model (MSM) by merging existing models of upper limbs and lower limbs~\cite{UpperLower5,UpperLower1,UpperLower3,UpperLower2,UpperLower4} and thoracolumbar spine~\cite{Spine1,Spine2, Spine3}. We add wrist and hand~\cite{Hand1,Hand2} models to the MSM, which are not used for ground truth generation, for~the sake of aesthetics. The~full-body model contains all bones in a skeletal system from the head to feet and from the upper arms to the hands. We do not model every degree of freedom between vertebrae to avoid expensive computation and the requirement of at least three markers to measure the motions of one vertebra. Instead, we separate the spine from the fifth lumbar to the first cervical vertebra into nine~segments. 

\xadded{Then, we fit our data to the musculoskeletal model. Instead of using the OMC marker data directly, we use OMC marker converted to 3D human mesh representations using the MoSh++ \cite{Mosh} method, to~make scaling the model to individual participants more time efficient and allow us to define an arbitrary number of virtual markers. We fit our data to the musculoskeletal model, by~first defining virtual markers on the vertices of the 3D mesh representation. We then used these virtual markers as input for the OpenSim software.
Then, we used the OpenSim internal scaling tool to scale the proportion of individual body segments according to the distances of virtual markers on the 3D mesh. As~the sizes of individual body parts vary across individuals, this step must be conducted individually for each subject in the dataset. We define the ratio in dimensions between the default and scaled body segments as scaling factors. Finally, we used the inverse kinematics solver for the calculation of joint angles. During~this process, the~MSM is moved for each time step to a position that minimizes the sum of weighted squared errors between the virtual markers on the 3D mesh and markers defined on the musculoskeletal model.}
All joint angles where segments had a higher squared error than \SI{2}{\centi\meter} were disregarded in the~analysis.

The final ground truth values were the calculated joint angles, the~scaling factors per segment as well as the virtual marker positions. Additionally, a~pelvis rotation matrix was generated 
% \jvg{motivate why}
for each frame, since the pelvis functions as the relative position of the model to the ground that is free to move in all~directions.

\subsubsection{Data Preparation and~Hyperparameters}

\xadded{To generate the input for our network, each video frame was cropped and augmented.}
We use the pre-trained Faster R-CNN~\cite{ren_faster_2016} with ResNet-50~\cite{resnet50} backbone to extract a square bounding box of the person in videos and resize it to $256 \times 256$ pixels as the input image size. During~training, we apply data augmentation with scaling, rotation, translation and noise to simulate occlusions similar to~\cite{MetricScale}.  

\xadded{Our model was trained using the following hyperparameters and loss.}
For the ResNeXt model, we use an Adam optimizer with weight decay~\cite{AdamW} of 0.001 and a batch size of 64. The~learning rate exponentially decays in two steps from $5 \times 10^{-4}$ to $3.33  \times 10^{-5}$ over 28 epochs and from $3.33  \times 10^{-6}$ to $10^{-6}$ over 2 epochs.
For both sequential and convolutional networks, we set the hyperparameters with $\lambda_1=1.0$, $\lambda_2=2.0$, $\lambda_3=0.1$ and $\lambda_4=0.06$ experimentally.
\xadded{Due to memory constraints, we do not train convolutional and sequential models simultaneously, but~in succession, by~first training the convolutional model and then refining predictions using the sequential model.}

\subsection{Software~Tools}

All training was conducted in python using the PyTorch library~\cite{paszke_pytorch_2019}. The~pre-trained ResNext and FasterRCNN networks were obtained from the torchvision library~\cite{Torchvision2021}. 
All code for training and generation of ground truth will be made available in a Github repository: {\url{https://github.com/bittnerma/Direct3DKinematicEstimation}. %MDPI: Footnote is not permitted in our journal. Please include this paragraph in the maintext.
%MDPI: Please provide the access date of the URL in the following format: "URL (accessed on Day Month Year)". %MBittner: This is our own repository, I have emphasized it by the colon, in case it is crucial the last access date was 22 December 2022 please uncomment the following line if necessary. 
%(accessed on 22 December 2022)
.}

\subsection{Data}

We trained and tested D3KE on \xdeleted{different subsets of} the BML-MoVi Database~\cite{ghorbani_movi_2021}. 
BML-Movi is an extensive motion capture and video dataset, it contains recordings of 90 actors that each perform 20 kinds of everyday movements as well as \xreplaced{a random one}{one random movement}.
Motions were captured using inertial measurement units as well as a Qualisys optical motion capture system and videos were recorded using two \xreplaced{computer-vision}{industrial Point Gray} cameras.\xdeleted{as well as two mobile phones}
For this study, we used recordings from the calibrated Point Gray cameras (PG1, PG2) during recording session F as the full set of optical markers was used during this session. In~accordance with the anatomical plane that each camera is viewing during the initial T-Pose of the participants, we will refer to the camera view captured by PG1 as the frontal- and PG2 as the sagittal camera view. For~the generation of ground truth virtual markers, we use the 3D mesh representations of the Qualisys data that is provided in the larger AMASS dataset~\cite{AMASS}.
For analysis of the data, we divided the BML-Movi database into 63 participants for training, 16~participants for the testing, and~three participants for validation. This is common practice in the supervised training of deep neural networks~\xadded{\cite{zhang_dive_2021}}.
At training time, the~validation set is used to evaluate the accuracy of kinematic estimation after each training iteration on a portion of the data the network does not have access to, to~prevent overfitting on the training~set.  

\section{Experiments}

\subsection{Experiment 1: Direct vs. Multi-Step~Estimation}

To evaluate the accuracy of our direct 3D kinematic estimation approach (D3KE) for joint angle estimation, we compare its performance against multiple versions of the multi-step~approach. 

\subsubsection{Experiment 1-A: 3D Pose Based Kinematic~Estimation}
\xadded{
We first compare our direct estimation of kinematics and a 3D pose estimation multi-step baseline.
To create a fair comparison between direct and multi-step estimations, we implement a custom multi-step approach (CMS) that is trained on the same data as our direct approach.
For the CMS, we combine a \textit{3D} human pose estimation method with subsequent musculoskeletal modeling in OpenSim. We modify the metric-scale heatmaps~\cite{MetricScale} of the convolutional network to predict marker positions and SMPL keypoint positions in the metric scale.
As for D3KE, we exploit a sequence network to re-fine marker positions at the target frame. More specifically, the~convolutional network initially infers marker positions under a calibration pose (T-pose), and~OpenSim utilizes the predicted marker data for body scaling, where the general musculoskeletal model is scaled to the participant's body size. Re-fined marker positions are then used to run inverse kinematics with the scaled musculoskeletal model to obtain joint angles. The~main difference between the CMS approach and D3KE is that CMS uses multiple steps to estimate the kinematics and is only supervised on the marker/pose estimation task, while D3KE is directly trained on the kinematic estimation task; this way, we can compare direct vs. multi-step estimation of kinematics.} \xadded{We use multiple metrics for the comparison of D3KE and the CMS. The~mean per bony landmarks position error (MPBLPE) is used to evaluate bony landmark positions. Bony landmarks are markers placed where bones are close to the surface, such as the elbow. This metric is inspired by the mean per joint position error (MPJPE) which is often used in 3D pose estimation. MPBLPE first aligns estimations and ground truth at the root position and calculates the average Euclidean distance. We directly evaluate the body scale factors by the root mean square error RMSE$_{body}$ on the scalars predicted by the network. However, to~present the results in a more intuitive format, we choose the axis along the longest dimension in each body scale and convert the scale of the axis into millimeters and calculate the mean absolute error (MAE$_{body}$). }

\subsubsection{Experiment 1-B: 2D-Pose Based Kinematic~Estimation}
\xadded{

In the previous experiment, we evaluate the multi-step baseline with the \textit{3D} body pose estimation method. However, the~use of fully trained \textit{2D} pose estimation algorithms is common in kinematic estimation works~\cite{needham_can_2021,pagnon_pose2sim_2021,pagnon_pose2sim_2022}.
Therefore, we conduct experiments to compare our method and these 2D-based kinematic estimation methods.
In contrast to our CMS method which estimates 3D pose from a single camera estimation, 2D pose estimation methods require at least 2 calibrated cameras for the estimation of 3D keypoints. The~use of an additional camera to generate the pose could be an advantage, which the CMS method does not have.  
We chose a naive implementation of the OpenPose algorithm~\cite{OpenPose,PytorchOpenPose2021}, which has extensively been used in related work~\cite{needham_can_2021,pagnon_pose2sim_2021,pagnon_pose2sim_2022}. Additionally, we test the MediaPipe implementation of the blazepose algorithm~\cite{bazarevsky_blazepose_2020}, as~a more modern 2D algorithm. MediaPipe is easy to use since it is available as a python library, however, in~contrast to OpenPose it runs faster, allows for additional smoothing of its predictions, provides more key points, and~is labeled on different keypoint labels. 
}

\xadded{
For the OpenPose and MediaPipe, we project the key points to 3D using the BML-Movi camera parameters {\url{https://github.com/saeed1262/MoVi-Toolbox} (accessed on 2 August 2022 %MDPI: Footnote is not permitted in our journal. Please include this paragraph in the maintext.
%MDPI: Please provide the access date of the URL in the following format: "URL (accessed on Day Month Year)". % MBittner: Done. Added 'accessed on 2 August 2022'
)}. For~OpenPose, we connected missing points (due to self-occlusion) using linear interpolation. For~MediaPipe, we chose the highest model complexity (2) and set enabled smooth landmarks for continuous frames of a video. 
For modeling and inverse kinematics, we follow the same steps as for the CMS method only redefining the positions of markers on the OpenSim model to fit the provided key points.
}

\xadded{
We compare against an average across both camera views for CMS and D3KE, as~MediaPipe and OpenPose need at least two cameras to work. We evaluate performance based on mean absolute error $MAE_{angle}(^{\circ})$, the~standard deviation of errors $SD_{angle}(^{\circ})$ and smoothness of the predictions as the mean velocity of the angle MV$_{angle}$ ($^{\circ}$/s).
The mean velocity error is calculated by the derivative of the landmark position and joint angle data with respect to time.}
\begin{equation}
\label{eq: velocity error}
    MV_{angle} = \sum_{t=0}^{n}  \frac{\frac{|s_t - s_{t+1}|}{\Delta t}}{n}
\end{equation}
with $s_t$ an individual marker position at time t, $\Delta t$ is the amount of time between time steps and n is the total number of~timesteps.

\subsection{Experiment 2: Sequential Network~Variants}
\xadded{
Since we have multiple options for the sequential networks, we evaluate three
% For our method, we investigate different sequential architectures 
to determine whether the additional modeling of temporal dependencies in the data improves the accuracy of our method or not.
For subsequent smoothing and reduction of self-occlusion artifacts of the estimations, we test three different networks including LSTM~\cite{LSTM}, temporal convolutional networks (TCNs) \cite{pavllo:videopose3d:2019}, and~a lifting Transformer~\cite{Transformer} as the sequential network. As~smoothing is known to improve the accuracy of multi-step approaches~\cite{needham_can_2021}, we also evaluate combinations of our CMS model with these sequential networks. 
}

\xadded{
For the LSTM, we implement a bidirectional architecture with a hidden size of 128, three recurrent layers and a dropout probability of 0.1. For~TCNs, we follow~\cite{pavllo:videopose3d:2019} to exploit 243 frames as the receptive field and make the momentum of batch normalization decay from 0.1 to 0.001. For~the lifting Transformer, we use a hidden size of 256 and 8 parallel attention heads in the self-attention layer and a channel size of 512 in the convolutional layer. Each sequential network is trained with a sequence length of 243 frames and a batch size of 128 over 50 epochs with Adam optimizer~\cite{Adam}. The~learning rate exponentially decays from $10^{-3}$ to $5 \times 10^{-6}$.  
}

\xadded{
We use the same metrics for the comparison of individual network variants as we used for the comparison of D3KE and CMS.
In addition, we investigate the smoothness of the predicted sequences, we estimate the mean velocity (MV) on bony landmark positions and joint angles, denoted as MV$_{BL}$ and MV$_{angle}$. 
}
\subsection{Experiment 3: Processing~Speed}
\xadded{
One important property of our proposed method for clinical applications is its processing speed.
As applications for camera-based kinematic estimation should form an alternative to visual examinations in the future, it should ideally be able to run fast enough to estimate kinematics from video frames as fast as they are collected by a camera, mostly between 15 and 30 frames per second.  
}

\xadded{
We compare the running time on Windows 10 with four core CPU, 52 GB RAM and NVIDIA T4 GPU. We compare the CMS and D3KE method as they both use the same type of convolutional network. We choose the lifting Transformer as the sequential architecture in both the CMS and D3KE. For~the CMS method, OpenSim is executed in parallel with four cores.
Our report results in frames per second. 
}
\subsection{Experiment 4: Generalization~Performance}

\xadded{

The goal of camera-based kinematic estimation is ultimately to create tools for researchers and clinicians to analyze and diagnose human movement, these tools should not discriminate between different subjects and movements. We analyze whether our method generalizes to different subjects, movements, and~joints.
}

\xadded{
To assess how well D3KE generalizes, we compare the estimates of the proposed method to the ground truth on the time series of each of the 16 participants in the test set with respect to the performed movement, the~joint, the~camera view and the individual participant. For~this test, we use the best performing model from experiment 1 with the lifting transformer.
}

\xadded{
For all time series of joint angles, mean absolute error (MAE) and Pearson’s correlation coefficient ($\rho$) were calculated between the estimation from D3KE and the ground truth.
}

\xadded{
Central tendencies in the data are reported as a median and interquartile range of MAE, RMSE and $\rho$, as~the data are not normally distributed, as~assessed through visual inspection and confirmed by the Shapiro-Wilk test. For~completion, mean and standard deviation are also reported.
The absolute values of $\rho$ were categorized as weak, moderate, strong and excellent for $\rho \leq 0.35$, $0.35 < \rho \leq 0.67$, $0.67< \rho \leq 0.90$ and $0.90 < \rho$, respectively~\cite{taylor_interpretation_1990}.  
}

\subsection{Software and~Tools}

\xadded{
All data analysis was conducted in python 3~\cite{Python3} using the pandas library to generate descriptive statistics, SciPy library for the calculation of MAE, RMSE and Pingouin library for the calculation of $\rho$. 
}
% \input{3_results}
%\input{4_results_n}
%% Missing in this section:
\section{Results}
\subsection{Direct vs. Multi-Step~Estimation}
\subsubsection{A: 3D Pose Based Kinematic~Estimation}
As shown in Table~\ref{tab:comparison}, D3KE has better performance than CMS methods in terms of joint angles and body scales, and~these two factors are the key to kinematic estimation. Significantly, D3KE reduces 37.4\% of errors on MAE$_{angle}$ when comparing the CMS method with the Transformer architecture. Although~the proposed method has a slightly larger MPBLPE than the CMS, this metric is not related to the kinematic estimation and is only used as one of the losses during training in the proposed method (e.g., MPBLBE is not minimized). This indicates a gain in accuracy when directly estimating kinematics from the video instead of the multi-step approach of first estimating pose and then estimating~kinematics.

\begin{table}[H]
%\centering
\caption{Comparison of bony landmarks position (MPBLPE), body scales (MAEbody) and joint angles (MAEangle) between the estimation of and the ground truth across all participants, movements, joints and camera views. For~the custom multi-step approach (CMS) as well as our proposed method, we compare convolutional networks with different temporal networks. All versions of the proposed method show superior performance for the prediction of body scales and joint angle estimation. All CMSs show superior performance in estimating marker positions. Each method group shows better performance for the task it was optimized for, highlighting the importance of direct~optimization. Bold numbers indicate the best performance.}
\label{tab:comparison}
\newcolumntype{C}{>{\centering\arraybackslash}X}
\begin{tabularx}{\textwidth}{cCCcC}
\toprule
                          &                       & \textbf{MPBLPE (mm)}    & \textbf{MAE$\boldsymbol{_{body}}$ (mm)} & \textbf{MAE$\boldsymbol{_{angle}}$ ($\boldsymbol{^\circ}$)}  \\ 
\midrule
\multirow{4}{*}{D3KE}     & Convolutional      & 37.78          & 6.07            & 3.58                    \\
                          & Conv.+ LSTM        & 37.61          & 5.97            & 3.57                    \\
                          & Conv.+ TCNs        & 38.06          & 5.93            & 3.54                    \\
                          & Conv.+ Transformer & \textbf{36.98} %MDPI: Please add explanation for the bold numbers. If the bold format is not necessary, please remove it %MBittner: Added explanation in the Figure caption.
 & \textbf{5.90}   & \textbf{3.54}           \\ 
\midrule
\multirow{4}{*}{CMS} & Convolutional      & 35.04          & 6.25            & 5.89                    \\
                          & Conv.+ LSTM        & \textbf{33.74} & -               & 5.79                    \\
                          & Conv.+ TCNs        & 34.52          & -               & 5.82                    \\
                          & Conv.+ Transformer & 34.00          & -               & \textbf{5.66}           \\
\bottomrule
\end{tabularx}
\end{table}

In Table~\ref{tab:bodyScales}, we list RMSE and MAE for body scales of selected segments. The~results show that the CMS performs better than D3KE on lower limbs, and~D3KE performs better than the CMS on upper limbs. The~CMS and the proposed method have comparable performance in scale estimation of the pelvis and lower~limbs.

\begin{table}[H]
%\begin{center}
\caption{Errors in scaling factors of the proposed method and the baseline compared against the ground truth. D3KE shows better performance for the upper extremities and slightly worse performance for the lower~extremities.  }
\label{tab:bodyScales}
\newcolumntype{C}{>{\centering\arraybackslash}X}
\begin{tabularx}{\textwidth}{CCC}
\toprule
  \multicolumn{3}{c}{\textbf{RMSE$\boldsymbol{_{body}}$ (MAE$\boldsymbol{_{body}}$ (mm))}} \\
\midrule
 &  \textbf{CMS} & \textbf{D3KE} \\
\midrule
pelvis & 0.090 (9.58) %MDPI: Please add explanation for the bold numbers. If the bold format is not necessary, please remove it %MBittner: I have removed them.
 & 0.091 (9.82) \\
femur &  0.073 (10.55) & 0.091 (22.21)\\
tibia &  0.060 (9.82) & 0.102 (35.00)\\
humerus & 0.102 (14.55) & 0.068 (9.41)\\
ulna &  0.395 (24.91) & 0.075 (11.59)\\
radius &  0.395 (23.60) & 0.075 (10.98)\\
\bottomrule
\end{tabularx}
%\end{center}
\end{table}
\unskip

\subsubsection{B: 2D Pose Based Kinematic~Estimation}

The results of the comparison of our proposed method, CMS, OpenPose, and~MediaPipe are shown in Table~\ref{tab:algorithm_comparison}. We find that algorithms trained on noisy labels, that use fewer key points perform worse than ours. We see a clear difference between the unsmoothed OpenPose estimations and the smoothed MediaPipe estimations in the mean velocity of the estimations. 
Our proposed method D3KE still performs better showing that even in an ideal scenario (CMS, i.e.,~no noise in the labels, enough markers, same distribution training data) direct estimation is~preferable. 

\begin{table}[H]
\caption{Comparison of popular pose estimation algorithms to D3KE. As~OpenPose and MediaPipe require multiple cameras to create 3D keypoints, we compare against the average of both camera views for CMS and D3KE. CMS shows better performance than OpenPose and MediaPipe and D3KE shows the overall best performance. Indicating that direct estimation is preferable to (naive) implementations of multi-step~methods.}
\label{tab:algorithm_comparison}
   \newcolumntype{C}{>{\centering\arraybackslash}X}
\begin{tabularx}{\textwidth}{CCCC}
\toprule
     & \textbf{$\boldsymbol{MAE_{angle}}$~($\boldsymbol{^\circ}$)} & \textbf{$\boldsymbol{SD_{angle}}$~($\boldsymbol{^\circ}$)} & $\boldsymbol{MV_{angle}}$ \textbf{($\boldsymbol{^{\circ}}$/s)}\\
    \midrule
    OpenPose & 16.98 & 25.91 & 75.15 \\
    MediaPipe & 10.60 & 18.80 & 37.15 \\
    CMS & 5.11 & 10.27 & 15.74\\
    D3KE & 3.41 & 6.05 & 13.57 \\
    \bottomrule
    \end{tabularx}
\end{table}
\unskip

\subsection{Sequential Network~Variants}
\xadded{
Table~\ref{tab:comparison} also shows the results of different sequential networks for smoothing of the predictions. Although~the convolutional model by itself already has good performance in joint angle and scale factor estimation, using temporal smoothing can additionally reduce the estimation error. 
}

The results of our investigation to reduce the noise in the estimations using temporal smoothing are shown in Table~\ref{tab:smoothness}. The~result shows that all temporal models can improve the smoothness of the sequence. The~LSTM achieves the best performance on MV$_{BL}$ and MV$_{angle}$ among all temporal models, this is contrary to the results in Table~\ref{tab:comparison}, in which using a Transformer as the sequential model yielded the best~results. 
\begin{table}[H]
%\begin{center}
\caption{The mean velocity errors for bony landmarks MV$_{BL}$ and joint angles MV$_{angle}$, lower values indicate smoother estimations. Adding a sequential model most probably improves the continuity of~estimations.}
\label{tab:smoothness}
\newcolumntype{C}{>{\centering\arraybackslash}X}
\begin{tabularx}{\textwidth}{CCC}
\toprule
%    \midrule
      &  \textbf{MV$\boldsymbol{_{BL}}$ (mm/s)} & \textbf{MV$\boldsymbol{_{angle}}$ ($\boldsymbol{^{\circ}}$/s)} \\
      \midrule
    Convolutional model & 378.01 & 21.7 \\
    Conv. + LSTM & 243.23 & 12.19 \\
    Conv. + TCNs & 245.35 & 12.29 \\
    Conv. + Transformer & 262.82 & 13.57 \\
    \bottomrule
    % OpenPose & 506.88 & 75.15 \\
    % MediaPipe & 453.58 & 37.15 
    \end{tabularx}
%\end{center}
\end{table}

\subsection{Processing~Speed}

Our proposed method achieves 31.96 fps with a batch size of 256, as shown in Table~\ref{tab:timeCost}. Since the \xadded{skeletal-model} layer must traverse body segments in the level order, our proposed method is slower than the CMS for a batch size of 1. However, the~support of mini-batch computation in the \xadded{skeletal-model} layer allows D3KE to run faster than the CMS. \xadded{Showing that our method can reach video framerate speeds} on a competent~GPU. 
\begin{table}[H]
%\begin{center}
\caption{Comparison of processing speed in FPS of D3KE and the baseline for multiple images or 'batches' in parallel. OpenSim does show little change in processing speed for increasing batch sizes. The~proposed method achieves framerate speeds for batches of 256 images, allowing it to analyze images as fast as a common webcam or mobile phone camera collects~them. Bold numbers indicate best performance.}
\label{tab:timeCost}
\newcolumntype{C}{>{\centering\arraybackslash}X}
\begin{tabularx}{\textwidth}{CCCCCC}
\toprule
\textbf{Batch Size} &  \textbf{1} & \textbf{16} & \textbf{64} & \textbf{128} & \textbf{256}\\
\midrule
D3KE & 0.92 & 8.78 & 20.94 & 28.25 & \textbf{31.96} %MDPI: Please add explanation for the bold numbers. If the bold format is not necessary, please remove it. %MBittner: I have added an explanation to the figure caption. 
\\
CMS &  7.51 & 8.36& 8.43& \textbf{8.44} & 8.35\\
\bottomrule
% OpenPose & 8.64 & 8.28 & - & - & -\\
% MediaPipe & 10.1  & - & - & - & - \\
% OpenPose & 2.85 & 2.81 & - & - & -\\
% MediaPipe & 3.00  & - & - & - & - \\
% \hline
\end{tabularx}
%\end{center}
\end{table}

\subsection{Generalization~Performance}

\begin{figure}[H]
    \includegraphics[width=\linewidth]{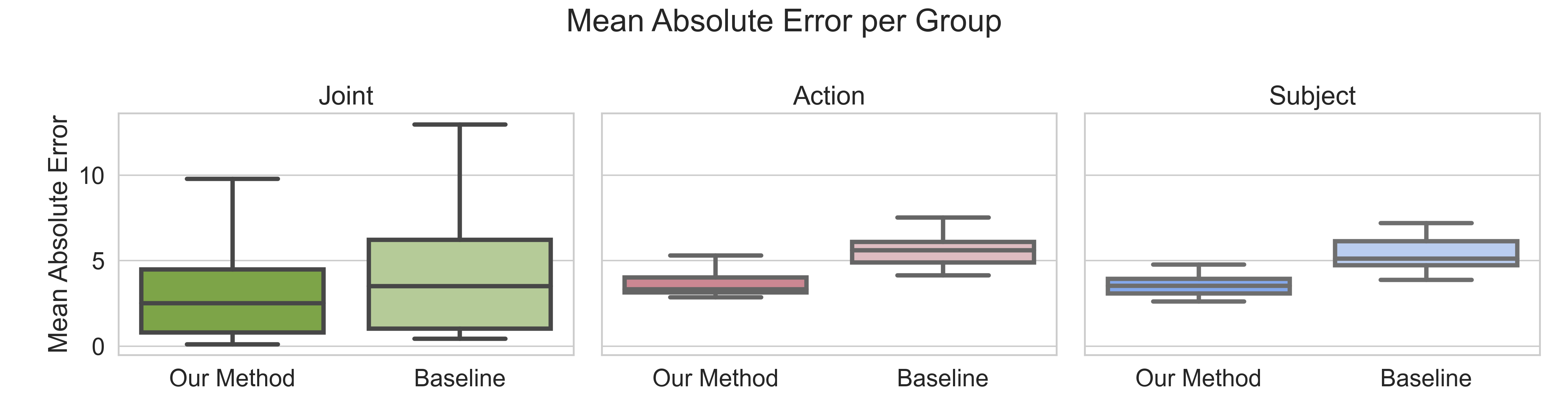}
    \caption{Mean absolute error for predicted joint angles per joint, movement and subset. Across these groups, D3KE shows less variation compared to the CMS. The~low variations indicate that D3KE is suitable for use on different participants and~movements.}
    \label{fig:mae_per_group}
\end{figure}

Figure~\ref{fig:mae_per_group} shows that both CMS and D3KE have relatively little variation across different movements and different participants, yet larger variations across individual joints. This is also reflected in median MAEs and $\rho$ per joint (Table \ref{tab:subgroup_results}), with~median MAEs for joints varying within a range \SI{3.8}{\degree} for joints, while movements and participants vary under \SI{1}{\degree}.
It shows that D3KE generalizes well to different participants and movements.
% Outliers of both methods can however occasionally reach \SI{180.63}{\degree} or perfect negative correlation with a $\rho$ of \SI{-1}{}.  

\vspace{-6pt}

\begin{table}[H]
    \caption{Median and inter-quartile ranges (IQR) of joint angles per \emph{joint}, \emph{movement} and \emph{participant}. MAE, RMSE and correlation are calculated over individual frames, Medians and IQR are reported due to the skewed distribution of results. Within~each group, both camera views show similar errors. Joint angles show the highest error and highest spread of values of all groupings. D3KE generalizes well to different movements, participants and camera~views.}
    \label{tab:subgroup_results}
\newcolumntype{C}{>{\centering\arraybackslash}X}
\begin{tabularx}{\textwidth}{ccCcCcCc}
\toprule
            \textbf{Group} &  \textbf{Camera View} & \multicolumn{2}{c}{\textbf{MAE ($\boldsymbol{^\circ}$)}} & \multicolumn{2}{c}{\textbf{RMSE ($\boldsymbol{^\circ}$)}} & \multicolumn{2}{l}{$\boldsymbol{\rho}$} \\
                & {} & \textbf{Median} &  \textbf{IQR} & \textbf{Median} &  \textbf{IQR} &      \textbf{Median} &  \textbf{IQR} \\
 %        \textbf{\hl{Group}} %MDPI: please check if the format of header correct? should it move to the above row? %MBittner: The position of the header was incorrect. I moved the header to the above row.
 % & \textbf{Camera View} &        &      &        &      &             &      \\
        \midrule
        Joint & Frontal &   2.13 & 3.80 &   2.54 & 3.96 &        0.77 & 0.16 \\
                & Sagittal &   2.14 & 3.03 &   2.55 & 3.49 &        0.73 & 0.21 \\
        Movement & Frontal &   1.85 & 0.63 &   2.19 & 0.84 &        0.76 & 0.11 \\
                & Sagittal &   1.91 & 0.46 &   2.30 & 0.65 &        0.74 & 0.11 \\
        Participant & Frontal &   1.76 & 0.53 &   2.03 & 0.66 &        0.77 & 0.04 \\
                & Sagittal &   1.84 & 0.28 &   2.14 & 0.35 &        0.74 & 0.04 \\
        \bottomrule
    \end{tabularx}
\end{table}
\unskip

\subsection{Qualitative~Results}
\begin{figure}[H]
%    \begin{center}
      \includegraphics[width=0.98\linewidth]{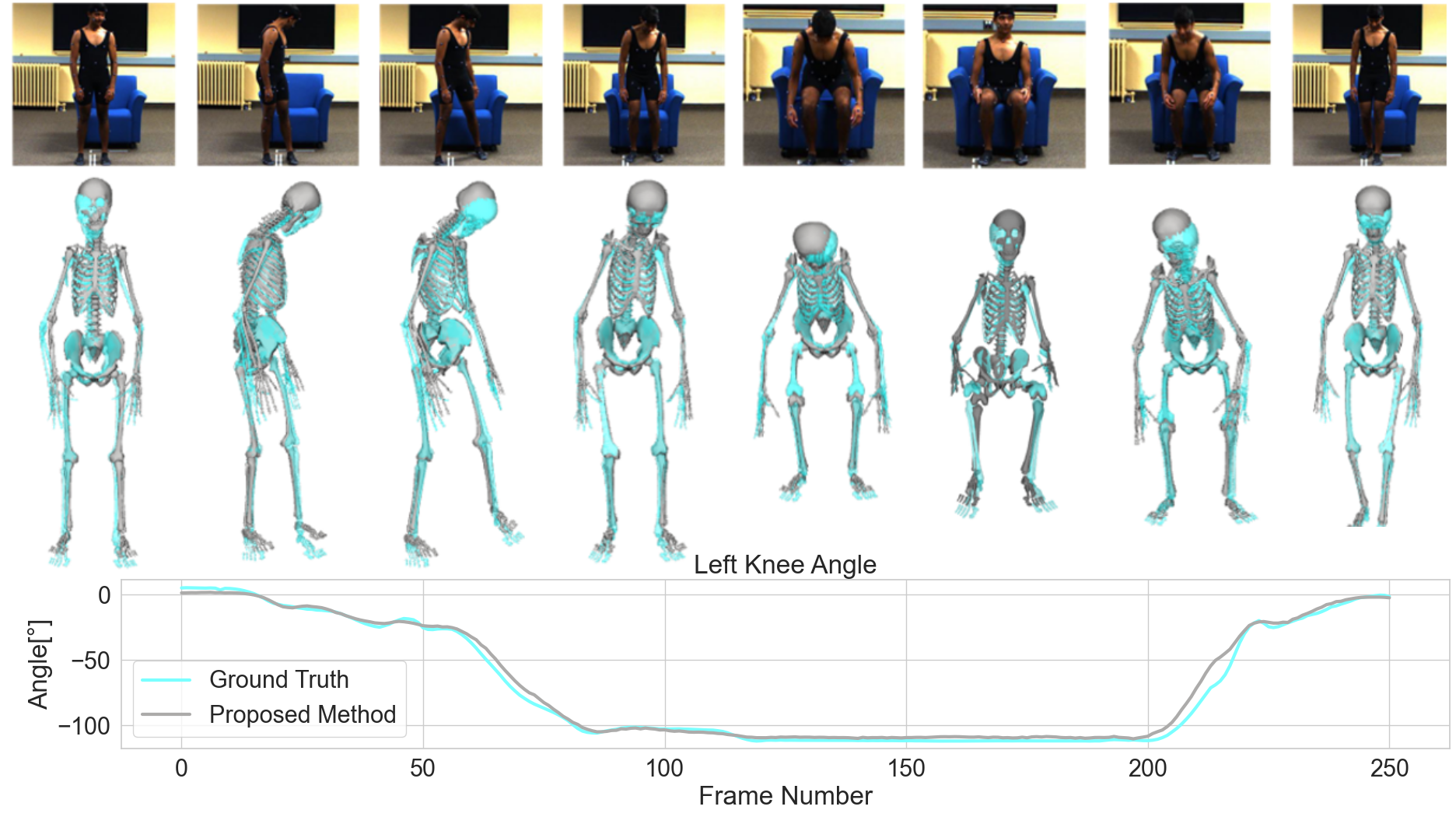}
%    \end{center}
      \caption{Qualitative results of D3KE from a 'sitting-down' movement in the BML-Movi dataset. The~top row shows selected frames throughout the movement. The~middle row shows different poses of the ground truth skeletal model throughout the movement (cyan) and the skeletal model (white) based on D3KE's estimation. The~bottom row shows the changes in flexion/extension of the left knee throughout the movement with blue being the predicted and orange being the ground truth~angle.  }
    \label{fig:qualitativeResults}
\end{figure}

We visualize the estimation of musculoskeletal models from our proposed method with the Transformer architecture in Figure~\ref{fig:qualitativeResults}. We also show the comparison between estimation and ground truth of the left knee angle as an example of joint angle estimation quality. We took the average body scales of the predicted sequence of scaling factors to scale the model and visualize it in OpenSim using the predicted joint angles as inputs. From~the figure, we can see that the proposed method can achieve results that are in agreement with the single-view input~video. 

\unskip

%\unskip
% \input{4_discussion}
%\input{5_discussion_n}
\section{Discussion}

In summary, we compared a direct approach of estimating joint angles from video images to the more traditional multi-step approach found in most recent works. The~traditional method first estimates key points from a video of a subject, then calculates joint angles using a (musculo)skeletal model through an inverse-kinematics process. We developed a method consisting of a convolutional neural network and a sequential network both including a specialized layer that performs kinematic transforms of a (musculo)skeletal model and allows for direct optimization of the predicted joint angles (D3KE) and treats the prediction of key points only as an auxiliary task. We compared our direct estimation approach against naive implementations of often used algorithms in the related literature\xadded{, as~well as a self-implemented custom multi-step approach (CMS) that is trained on the same data as our direct approach. We show that direct estimation of kinematics yields higher accuracy in predicted joint angles compared to the traditional multi-step approach. Our results indicate that direct estimation can help the future development of algorithms for fast and accessible kinematic analysis for researchers and clinicians.}

\subsection{Direct vs. Multi-Step~Estimation}
\subsubsection{3D-Pose Kinematic~Estimation}

\xadded{To compare direct estimation vs. multi-step estimation, we compared our D3KE method against a 3D-pose based multi-step approach (CMS) with comparable network architecture and trained on the same training data.} Compared to the CMS, our proposed method improves the accuracy of joint angle estimation. For~all model combinations, we can see an improvement of about $35\%$ in accuracy for the estimation of joint angles. Our results support the feasibility of our proposed method. It delivers improvements due to directly optimizing the predicted joint angles and scaling factors while using the pose estimates only as an auxiliary task. The~auxiliary task effectively imposes a constraint on the network estimation. We show that direct optimization is preferable to the multi-step approach when using videos from a single-camera view. We expect that using additional specialized layers, a network might be able to directly optimize for individual muscle forces with comparable accuracy from a monocular~video. 

\subsubsection{2D-Pose Kinematic~Estimation}

\begin{figure}[H]
%    \begin{center}
      \includegraphics[width=1.0\linewidth]{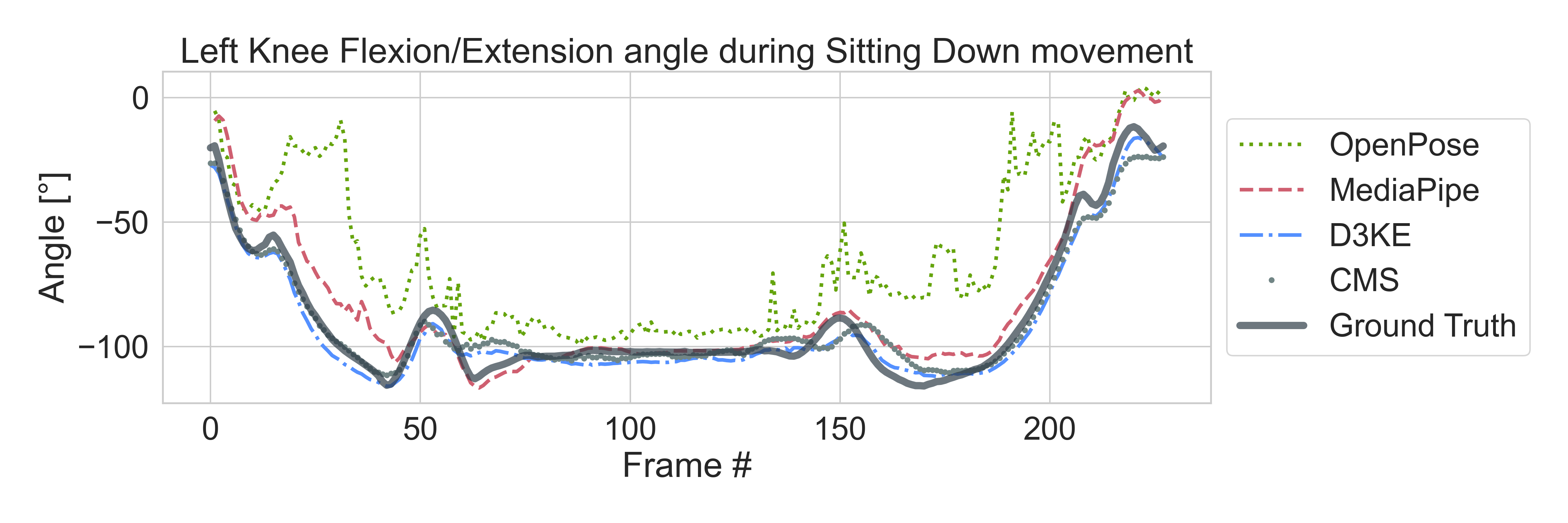}
%    \end{center}
      \caption{Qualitative comparison of predicted angles of the left knee, from~a 'sitting-down' movement in the BML-Movi dataset. While OpenPose shows by far the noisiest estimation, the~smoothing of the MediaPipe estimation is clear, our proposed method and implemented CMS work best, probably due to the restriction of additional~markers. }
    \label{fig:algorithm_lineplots}
\end{figure}

\xadded{We compared our direct approach (D3KE) and the self-trained multi-step approach (CMS) against two multi-step approaches commonly used in the literature.}
Compared to the more traditional implementations of the OpenPose and MediaPipe algorithms, both our proposed and our CMS method show superior performance. From~Tables~\ref{tab:smoothness} and \ref{fig:algorithm_lineplots}, we see that estimations from D3KE are far smoother compared to the traditional methods. This is likely due to multiple reasons. 
\xadded{The predicted key points of OpenPose suffer from systematic errors due to inaccuracies in their training data~\cite{cronin_using_2021}, which can explain the drop in performance, for~MediaPipe the accuracy of labels in their training data is not known as their paper only states that their annotators were human~\cite{bazarevsky_blazepose_2020}, not whether they had expertise in labeling anatomical key points.
The lack of smoothing for the predicted OpenPose key points can also contribute to its overall worse performance. MediaPipe, which uses internal smoothing, shows better results in comparison.}
In general, we expected worse performance from OpenPose and MediaPipe as they only predict 18 or 33 key points respectively, while we supervise a total of 77. Both traditional methods predict keypoints representing joint centers and not markers on body segments, this makes it hard to distinguish rotations between different body segments during the musculoskeletal modeling~step.

\subsection{Network~Variants}

\xadded{We compared three commonly used sequential networks to improve our estimation. We show that all sequential networks improve our estimation accuracy. One possible explanation for this is that the network is better able to handle `self-occlusion' artifacts. The~estimation of a 3D-pose from a single camera is an ill-posed problem as a 3D point projected to a 2D image can originate from any position along the ray(s) that fall on the image sensor and form the corresponding pixel. This can lead to self-occlusion, such as the torso and left arm occluding the right arm during a right arm swing in frames recorded from the left sagittal view. During~self-occlusion, it is difficult for frame-based networks to make a good estimate as they lack temporal information of previous angles of the arm to extrapolate from. Sequential networks on the other hand have access to temporal information, which can allow for more accurate estimations. Although~we only see a slight increase \SI{0.001}{\degree} in MAE of the joint angle estimation, we can see a clear improvement of the sequential models in the smoothness of the predicted angles Table~\ref{tab:smoothness}. 
This might be due to the network learning to interpolate motions during occurrences of self-occlusion.
}
\subsection{Processing~Speed}

\xadded{Compared to the multi-step baseline, CMS, our D3KE approach shows increased calculation speeds for larger batch sizes. Both CMS and D3KE make use of the same ResNeXt50 architecture, which should show approximately the same performance increase with increasing batch sizes for both methods. D3KE could be expected to be slower, as~it also has the additional time cost of calculating the pose from the estimated kinematics in the skeletal model layer. However, due to its multi-step nature, CMS has to perform an additional inverse kinematics calculation. This calculation seems to form a bottleneck in the processing speed of the CMS approach restricting it to a framerate of ~8 fps. Other multi-step algorithms will most likely encounter the same problem. In~the case of OpenPose, which runs at about 4 fps~\cite{bazarevsky_blazepose_2020}, even lower frame rates can be expected for a complete pipeline. This shows the advantage in the processing time of our direct approach. }

For a method to be usable in everyday life, it should be reasonably fast in running. Processing speeds allowing a method to run between 15 and 30 frames per second \xadded{are favorable, as they show that a method can process a video as fast as its frames are collected.
However, our results might not directly translate to every real-world scenario. To~process multiple images simultaneously as batches, D3KE currently requires GPUs that are not available in mobile devices, which prevents it from being%Please verify.
portable. 
In addition, we use the Faster R-CNN object detection network to crop our images. This step was not included in the processing speed evaluation, as it is highly dependent on the chosen object-detection algorithm. However, with~inference speeds of ~12fps, the Faster R-CNN object detection would form a bottleneck in applying our method in real-time applications. Given the speed of development in the field of object detection, Faster R-CNN can by now be regarded as an old algorithm, and newer and faster object detectors should be used instead. The~YoloV7 algorithm~\cite{wang_yolov7_2022}, which performs object detection at up to 286 fps could be considered.}
\xadded{In general,} the current architecture is not optimized for speed or a specific technology and we are using off-the-shelf, fairly standard convolutional and sequential architectures. For~these architectures, smaller and faster alternatives might be found in the future. 
When optimizing for all these points, we predict the proposed method could run on mobile devices within a few years, effectively enabling a 3D kinematic analysis instrument to become available for everyone with a mobile phone or~tablet. 

\subsection{Generalization~Performance}

\xadded{Our method generalizes well on the tested data.} As shown in Figure~\ref{fig:mae_per_group} and Table~\ref{tab:subgroup_results}, the~estimation variations across participants and movements are small. We can conclude that our method \xadded{shows the ability to generalize} to different camera views, participants, and~performed movements, within~the tested dataset. Our results indicate that D3KE could be generally applied to a variety of people and movements, including clinical and sports applications, e.g.,~physiotherapists and athletes, \xadded{when trained on sufficient additional data.}

\xadded{Although we show good generalization performance on the BMLMovi database, it is difficult to estimate how well our method will generalize in a real-world scenario. In~machine learning settings, training data is often not representative of the task of the network in the real world~\cite{attenberg_beat_2015} and can introduce biases if applied to scenarios that are very different from the one represented in the training data. Unfortunately, there is currently a lack of deep learning datasets for kinematic analysis~\cite{seethapathi_movement_2019,needham_can_2021,cronin_using_2021,wade_applications_2022}. 
In addition, while the BML-Movi database is excellent for training neural networks due to the large number of participants performing movements and the diversity of execution styles, it might be not extensive enough to train a network for biomedical applications in the real world. However, to~evaluate the current method fully, such an extensive dataset would be necessary. In~general, we expect a drop in accuracy when our method is applied to a scenario different from the BMLMovi database. As~we train on just two calibrated cameras, we expect our method to be most vulnerable to alternative camera positions, that do not show people in either frontal or sagittal view. Future research should investigate the stability of direct estimation methods when applied to data that differs significantly from the training data.}

\subsection{Future~Work}

\xadded{To improve the accuracy of the algorithm and provide further insight into the strengths and weaknesses of monocular joint angle estimation,} a new dataset with dedicated annotation is needed. A~dataset specifically designed for the estimation of joint angles and/or kinetics could improve the accuracy of the algorithm.
This dataset could be established with a large number of camera views, and~top-down views for better estimation of movements in the transverse plane, where participants perform movements that exercise the full ROMs of individual joints including upper extremities, as~well as movements that are relevant for health care professionals such as physiotherapy exercises and other clinical tests. In~addition, the~inclusion of abnormal movement patterns could give better insights into the clinical relevance of newly developed~methods. 

\xadded{Transfer learning could be explored to apply 3DKE in settings where little training data are available. 
Vdeos that are very different from the BMLMovi training data, such as people wearing more clothes, are in different surroundings, or~are filmed from a different camera view, will, most probably, yield worse accuracy than shown in this paper.  
Transfer learning of a pre-trained D3KE on a minimal portion of a dataset could be investigated as an alternative to the time-consuming collection of a novel dataset.}

\xadded{The capabilities of D3KE as an adapter for kinetic analysis of a movement in OpenSim could be explored.  Given data similar to BMLMovi or successful transfer learning on relevant data beforehand, our method provides an easy way to skip the tedious steps of scaling and running inverse kinematics on an MSM.} This enables the quick generation of MSMs for kinetic analysis from just a single video. Even if this kinematic estimation comes at the cost of reduced accuracy, it could provide coarse insights into collected data, which can later be confirmed through finer analysis with the manually scaled~MSMs.

\xadded{D3KE could be made more generally applicable if the underlying model of the Skeletal-model layer would not be fixed. Currently, the~underlying model is fixed in the Skeletal-model layer. Future iterations could explore combinations of the Pytorch and OpenSim python libraries to allow training a network on a self-defined model or allowing a pre-trained model to be refined through transfer learning for, e.g.,~only estimation of joint angles around the shoulder.}

\xadded{Existing Explainable AI tools should be applied to better understand the inner workings of D3KE. 
Deep neural networks are capable of high accuracy estimation, because~of their ability to break down highly complex tasks into simpler tasks~\cite{allen-zhu_backward_2021}, but~understanding what these simpler tasks are is non-trivial. Research in Explainable AI has generated tools and frameworks that allow one to better understand the basis of the final predictions of a network~\cite{selvaraju_grad-cam_2017}. 
Applying these tools could help users and researchers alike to better understand the biases and limitations of our method. D3KE can still predict the joint angles even if these joints are occluded; this means it must make assumptions. What these assumptions are and how they came to be are important to estimate the trustworthiness of this algorithm in a real-world scenario.}

\section{Conclusions}

In this paper, we present a novel end-to-end neural network for the estimation of segment joint angles of the human body. Compared to the previous method, we directly regress to the joint angle and scale for individual segments from the input video.  We trained our method from scratch on the BML-Movie database and compared it against a 3D pose estimation method on which we used the inverse kinematics tool of OpenSim to obtain the~kinematics.

We conclude that using direct estimation of joint angles is preferable in a single camera setting, as it is more accurate compared to the common approach of fitting an estimated pose to a musculoskeletal model and performing inverse kinematics. By~allowing the network to directly optimize for the joint angles and scaling factors, our method is less prone to errors in the key point labels used to predict key point location for pose estimation.
In addition, the~use of a sequential model is important when designing a neural network architecture for kinematic estimation, as~it allows to smooth predictions over time to create better estimates of limb position and joint angles during self-occlusion. 
% The proposed method achieves a better result than the CMS in terms of the joint angle and scaling which are keys for the OpenSim model fitting. Additionally, we tested the influence of different sequential models to improve the smoothness of our predictions and explored the main source of errors for our best performing model. 

While using deep learning for biomedical solutions is still in its infancy, the~presented method shows that training networks from scratch for specialized tasks is a viable way to estimate joint angles from a single camera video. With~further advancements in the underlying algorithms as well computational  performance, we predict that the methodology we have presented will assist biomedical and clinic practitioners to measure and monitor human movement in the near~future.

% \section{Patents}

% This section is not mandatory, but may be added if there are patents resulting from the work reported in this manuscript.

%%%%%%%%%%%%%%%%%%%%%%%%%%%%%%%%%%%%%%%%%%
\vspace{6pt}

\paragraph{\textbf{Funding:} This work %MDPI: Information regarding the funder and the funding number should be provided. Please check the accuracy of funding data and any other information carefully. %MBittner: Funding information is correct, this is the same funding source as used for an earlier publication see https://www.mdpi.com/2076-3417/10/23/8630
 was supported by the Dutch Research Council (NWO) under the Citius Altius Sanius Perspective Program P16-28 Project 4.}

\paragraph{\textbf{Aknowledgements:} The authors would like to thank Lisa Noteboom for her feedback as well as Marco Hoozemans and Dirkjan Veeger for guidance and insight during our bi-weekly~meetings.}

% \conflictsofinterest{Declare conflicts of interest or state ``The authors declare no conflict of interest.'' Authors must identify and declare any personal circumstances or interest that may be perceived as inappropriately influencing the representation or interpretation of reported research results. Any role of the funders in the design of the study; in the collection, analyses or interpretation of data; in the writing of the manuscript; or in the decision to publish the results must be declared in this section. If there is no role, please state ``The funders had no role in the design of the study; in the collection, analyses, or interpretation of data; in the writing of the manuscript; or in the decision to publish the~results''.} 

\paragraph{\textbf{Conflict of Interest:} The authors declare no conflict of~interest.}
% %%%%%%%%%%%%%%%%%%%%%%%%%%%%%%%%%%%%%%%%%%
% %% Optional
% \sampleavailability{Samples of the compounds ... are available from the authors.}

% %% Only for journal Encyclopedia
% %\entrylink{The Link to this entry published on the encyclopedia platform.}

% \abbreviations{Abbreviations}{
% The following abbreviations are used in this manuscript:\\

% \noindent 
% \begin{tabular}{@{}ll}
% MDPI & Multidisciplinary Digital Publishing Institute\\
% DOAJ & Directory of open access journals\\
% TLA & Three letter acronym\\
% LD & Linear dichroism
% \end{tabular}
% }

\paragraph{\textbf{Abbreviations:} The following abbreviations are used in this manuscript:}

\begin{abbreviations}
    \item[CMS] Custom multi-stage approach
    \item[D3KE] Direct 3D kinematic estimation
    \item[IQR] Interquartile Range
    \item[MAE] Mean absolute error
    \item[MMC] Markerless motion capture
    \item[MPBLPE] Mean per bony landmark error
    \item[MSM] Musculoskeletal model
    \item[MVE] Mean velocity error
    \item[OMC] Optical Motion Capture
    \item[PCC] Pearson correlation coefficient
    \item[RMS] Root mean square error
    \item[ROM] Range of motion
    \item[SD] Standard Deviation
    \item[TCN] Temporal Convolutional Network
\end{abbreviations}

\printbibliography

\end{document}